\documentclass[runningheads]{llncs}
\usepackage[T1]{fontenc}
\usepackage{graphicx}
\usepackage{booktabs}
\usepackage[misc]{ifsym}

\graphicspath{{figures/}}
\usepackage{multirow}
\usepackage{hyperref}
\usepackage{color}
\usepackage{amsmath}
\usepackage{amssymb}
\usepackage{algorithm}
\usepackage{algpseudocode}
\usepackage{xcolor}
\usepackage{colortbl}
\usepackage{array}

\usepackage{mdframed}
\newenvironment{boxedrq}
  {\begin{mdframed}[roundcorner=2mm, linewidth=0.5pt, backgroundcolor=gray!5]
   \par\noindent\textbf{Main Research Question. }\itshape\ignorespaces}
  {\end{mdframed}}

\definecolor{rowbase}{HTML}{FDF3DA}
\definecolor{rowours}{HTML}{DCE8F5}
\makeatletter
\newcommand{\HlBase}[1]{%
  \State \hspace*{-\ALG@thistlm}%
  {\setlength{\fboxsep}{0pt}%
   \colorbox{rowbase}{\parbox[c][\baselineskip][c]{\linewidth}{%
     \hspace*{\ALG@thistlm}#1}}}%
}
\newcommand{\HlOurs}[1]{%
  \State \hspace*{-\ALG@thistlm}%
  {\setlength{\fboxsep}{0pt}%
   \colorbox{rowours}{\parbox[c][\baselineskip][c]{\linewidth}{%
     \hspace*{\ALG@thistlm}#1}}}%
}
\makeatother

\newcommand{\PhaseI}{\mbox{\texttt{Phase I}}}
\newcommand{\PhaseII}{\mbox{\texttt{Phase II}}}

\algnewcommand\algorithmicforeach{\textbf{for each}}
\algdef{S}[FOR]{ForEach}[1]{\algorithmicforeach\ #1\ \textbf{do}}

\begin{document}
\title{Federated Attack Campaign Detection\\via Contrastive Encoding of Threat Indicators in Gradient Updates}
\titlerunning{\textsc{FedIoC}: Attack Campaign Detection via IoC Encoding}

\author{%
Manuel Röder\inst{1,2} \and 
Bibin Babu\inst{1,3} \and
Frank-Michael Schleif\inst{1}
}%

\institute{
Technical University of Applied Sciences Würzburg-Schweinfurt, \\Würzburg, Germany,\\
\and
Bielefeld University, Bielefeld, Germany\\
\and
Center for Cybersecurity TTZ-WUE, Ochsenfurt, Germany
}%
\authorrunning{Röder et al.}

\maketitle

\begin{abstract}
Detecting orchestrated cyberattack campaigns that span multiple organizations traditionally requires sharing sensitive telemetry and threat intelligence across institutional boundaries and country borders, a barrier that Federated Learning removes by training shared threat detectors directly on local data.
We propose \textsc{FedIoC}, a modular framework in which clients fold locally available structured threat indicators into their gradient updates; we instantiate the client-side encoder with a supervised contrastive loss over IoC-matched flows.
Within each training batch, flows that match any known indicator pattern form the positive set; the contrastive objective pulls their learned embeddings together and pushes non-IoC embeddings away, so that campaign-relevant structure is, by design, expressed in the gradient direction.
Clients sharing indicators for the same attack campaign then produce aligned gradient components, which the server clusters by the cosine similarity of their updates to recover global campaign patterns without any direct IoC transmission.
We evaluate \textsc{FedIoC} on two public threat-detection benchmarks distributed across FL clients that each observe only a fragment of every active campaign and hold disjoint indicator sets derived from their local telemetry.
In this regime the FL server recovers cross-organizational campaign cohorts directly from gradient geometry.
We contribute \textsc{FedIoC} as a modular framework for this setting, and use it to pinpoint the non-IID gradient structure as the main driver of recovery and to define the open problem of designing encoders that improve on it.


\vspace{1.0em}
\noindent
\textbf{Reproducibility:} Code and setup instructions to reproduce the experimental results are available at \href{https://github.com/ManuelRoeder/fedioc}{https://github.com/ManuelRoeder/fedioc}.
\end{abstract}

\section{Introduction}
\label{sec:intro}

Detecting remote-orchestrated cyber attacks that span multiple organizations is fundamentally a fragmentation problem: no single defender observes the full set of Indicators of Compromise~(\textbf{IoC}) associated with a given campaign.
These artifacts (file hashes, IP addresses, domain names, URLs, TLS fingerprints, registry keys, and behavioral patterns) identify adversary tools, infrastructure, and tradecraft, but each organization holds only a partial view derived from its local telemetry.
Pooling indicators through threat intelligence sharing standards such as MISP~\cite{wagner2016misp} or STIX/TAXII\footnote{STIX (Structured Threat Information eXpression) is a standardized language for representing cyber threat intelligence; TAXII (Trusted Automated eXchange of Intelligence Information) is the corresponding transport protocol for sharing STIX content.} is the conventional remedy, yet its effectiveness is limited by sharing reluctance, data sovereignty constraints under regimes such as the GDPR~\cite{gdpr2016} and the NIS-2 directive~\cite{nis2directive}, and the rapid staleness of low-level indicators as adversaries rotate infrastructure and re-tool~\cite{alaeifar2024cti}.

Federated Learning~(\textbf{FL}) offers an alternative solution: organizations exchange gradient updates instead of raw telemetry~\cite{mcmahan2017fedavg}.
Existing FL systems for intrusion and threat detection nevertheless reduce IoC to label assignment~\cite{tabrizchi2025fedcyber,campos2022fl}: once a round begins, the indicators contribute nothing beyond the per-flow class label, and the resulting gradients carry only task-specific learning.
The richer STIX metadata that characterizes an IoC, comprising confidence, temporal validity, and kill-chain context, is therefore discarded before aggregation, and with it the campaign-cohort signal that gradient sharing could in principle convey.
We consequently investigate whether this signal can be preserved by asking:
\begin{boxedrq}
\label{ref:rq}
Can IoC knowledge be encoded into federated gradient updates, such that the FL server is able to resolve client-local IoC views into global attack campaign signals?
\end{boxedrq}
In our work, we aim to answer this question by proposing \textsc{FedIoC}, a modular FL framework which adds a supervised contrastive loss over indicator-matched flows to local client training.
Subsequently, the server is able to recover campaign structure by clustering clients on the cosine similarity of their gradient updates.
\begin{figure}[!t]
        \centering
    \includegraphics[width=\linewidth]{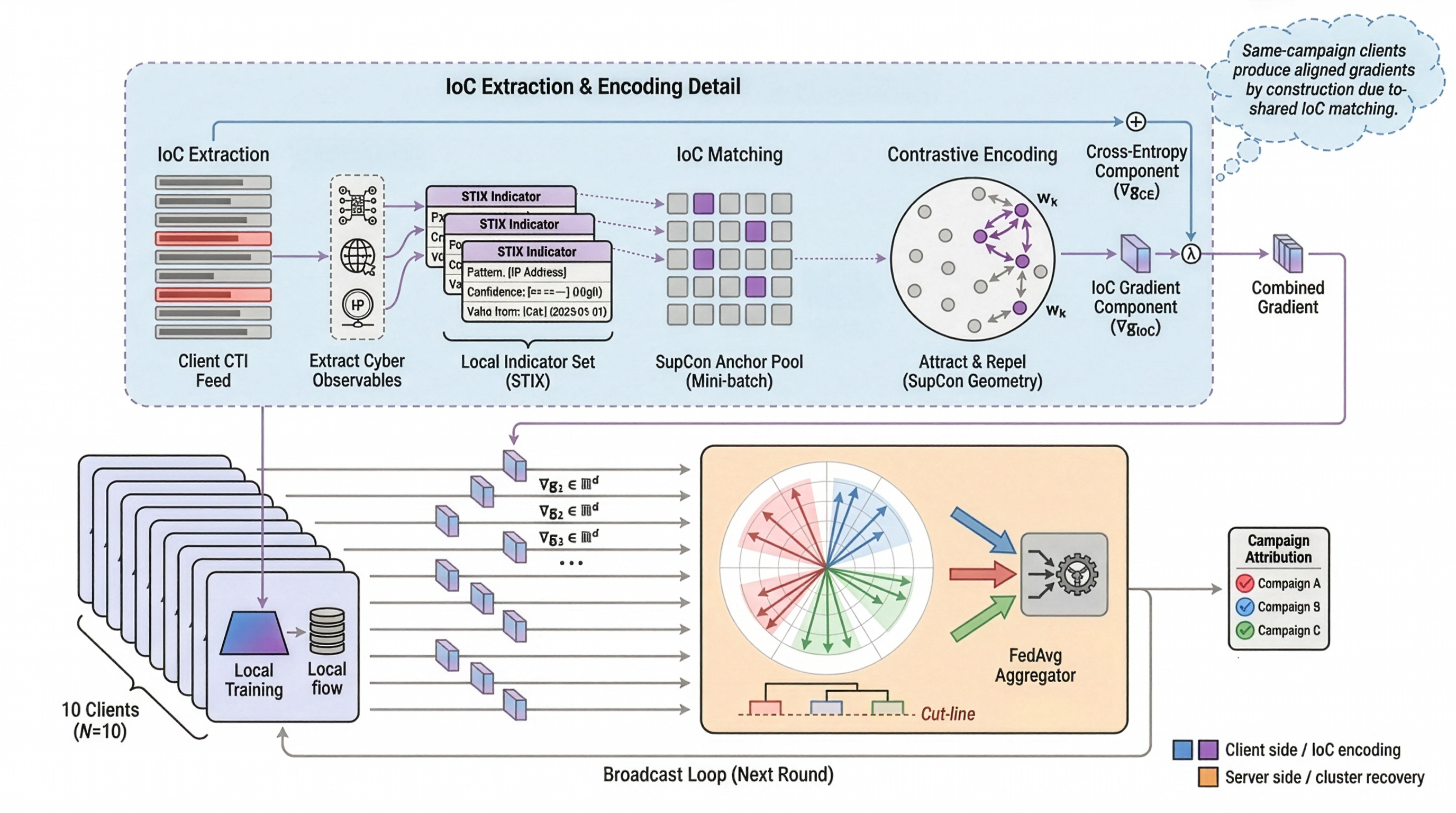}
    \caption{\textsc{FedIoC} pipeline overview.
    \emph{Client side (blue):} each client matches local mini-batch flows against its CTI feed and runs an indicator-weighted supervised contrastive pass alongside cross-entropy; the two pseudo-gradients sum into a single update that encodes campaign identity by design.
    \emph{Server side (orange):} cosine clustering over per-client gradient directions yields a campaign-cohort report (clients with shared exposure to the same attack infrastructure) before standard aggregation and broadcast.}
    \label{fig:overview}
\end{figure}

\noindent
\textbf{The key contributions are threefold:}
\begin{itemize}
    \item \textsc{FedIoC}, \emph{a modular end-to-end framework} (Sec.~\ref{sec:method}): clients encode locally available threat indicators into their gradient update, the server clusters the uploaded gradients by cosine similarity to recover campaign cohorts, and no raw indicator is transmitted. The client-side gradient encoder is interchangeable.
    \item \emph{Contrastive gradient encoding component} (Sec.~\ref{subsec:weighting}): a supervised contrastive loss over IoC-matched flows that we instantiate and evaluate; alternative encoders are left open for future work.
    \item \emph{Controlled empirical study on NIDS}\footnote{NIDS: Network Intrusion Detection Systems} \emph{benchmarks} (Sec.~\ref{sec:experiments}): We evaluate server-side campaign recovery and client-side detection on CTU-13 and UNSW-NB15 datasets.
\end{itemize}

\section{Methodology}
\label{sec:method}
We first formalize the federated attack campaign detection setting (Sec.~\ref{subsec:formulation}), then describe how each client identifies IoC-matched flows (Sec.~\ref{subsec:ioc_match}) and trains an indicator-weighted supervised contrastive objective that aims to imprint campaign identity onto the gradient direction (Sec.~\ref{subsec:weighting}); the server recovers the global campaign partition by clustering client gradients under cosine similarity (Sec.~\ref{subsec:server}), and we close by discussing compatibility with gradient-transmitting FL aggregators (Sec.~\ref{subsec:compat}). Figure~\ref{fig:overview} presents the overall pipeline of \textsc{FedIoC}.

\subsection{Problem Formulation}
\label{subsec:formulation}

Let $\mathcal{C} = \{c_1, \ldots, c_n\}$ be a set of $n$ federated clients, each holding a local dataset $\mathcal{D}_i$ of network traffic samples $(x, y)$ with binary or multi-class labels.
Each client $c_i$ has access to a set of indicator objects $\mathcal{I}_i^{(t)}$ extracted from a CTI\footnote{CTI: Cyber Threat Intelligence} feed at round $t$, where each indicator $\iota_k \in \mathcal{I}_i^{(t)}$ carries a pattern $\pi_k$ and a confidence score $\beta_k \in [0,1]$.
The pattern $\pi_k$ is treated abstractly as a predicate that decides whether a sample $x \in \mathcal{D}_i$ matches the indicator.
\textsc{FedIoC} uses the pattern field $\pi_k$ to define the IoC-matched anchor set $\mathcal{X}_i^{\mathrm{IoC}}$ (see Eq.~\eqref{eq:ioc_match}) and the confidence score $\beta_k$ as the per-anchor contrastive weight $w_a$ in the loss (Sec.~\ref{subsec:weighting}).
Further, let $\mathcal{K} = \{1, \ldots, K\}$ be the set of ground-truth attack campaigns.
Each indicator $\iota_k$ belongs to exactly one campaign $\kappa(\iota_k) \in \mathcal{K}$.
No single client holds all indicators for any campaign: each set $\mathcal{I}_i^{(t)}$ is a strict subset of the full collection of indicators for any campaign represented in client $c_i$'s traffic.

\textbf{Goal.}
Design a client-local training objective $\mathcal{L}_i(\theta)$ such that:
(1)~the gradient $g_i^{(t)} = \nabla_\theta \mathcal{L}_i(\theta^{(t-1)})$ encodes the IoC knowledge available to client $c_i$ at round $t$;
(2)~clients whose $\mathcal{I}_i^{(t)}$ share campaign membership produce gradients that are similar in cosine distance;
(3)~a server-side clustering of $\{g_i^{(t)}\}_{i=1}^n$ recovers the campaign partition $\kappa$, without any client transmitting raw telemetry data.

\subsection{IoC-Matched Sample Identification}
\label{subsec:ioc_match}

At each round $t$, client $c_i$ identifies the subset of its local training flows that match the indicators currently available to it:
\begin{equation}
    \mathcal{X}_i^{\mathrm{IoC}} = \bigl\{ x \in \mathcal{D}_i \;\big|\; \exists\, \iota_k \in \mathcal{I}_i^{(t)} : x \models \pi_k \bigr\},
    \label{eq:ioc_match}
\end{equation}
where $\pi_k$ is the predicate associated with indicator $\iota_k$ and $x \models \pi_k$ denotes that $x$ satisfies it.
We suppress the round superscript $(t)$ on $\mathcal{X}_i^{\mathrm{IoC}}$ for readability; it is understood to depend on $\mathcal{I}_i^{(t)}$.
\textsc{FedIoC} treats $\pi_k$ abstractly: any matching rule a client can evaluate on a local flow is admissible.
STIX cyber observables~\cite{stix21} are one concrete instantiation that supply both the predicate and the confidence score used as the per-anchor weight $w_a$.

\subsection{Contrastive IoC Objective}
\label{subsec:weighting}

For a mini-batch $\mathcal{B}$, let $\mathcal{A}(\mathcal{B}) = \{b \in \mathcal{B} : x_b \in \mathcal{X}_i^{\mathrm{IoC}}\}$ be the IoC-matched anchors and $P(a) = \mathcal{A}(\mathcal{B}) \setminus \{a\}$ the positives of anchor $a$.
Let $z_b = \phi_\theta(x_b) / \|\phi_\theta(x_b)\|_2$ be the $\ell_2$-normalized penultimate embedding (so $z_a \cdot z_p = \cos\bigl(\phi_\theta(x_a), \phi_\theta(x_p)\bigr)$).
For each anchor $a$ matched by indicator $\iota_{k(a)}\in\mathcal{I}_i^{(t)}$, we define the indicator-derived weight $w_a := \beta_{k(a)} \in [0,1]$, where $\beta_{k(a)}$ is the confidence score of the matching indicator. 
Subsequently, the IoC contrastive loss is an indicator-weighted variant of the Supervised Contrastive Learning (\textbf{SupCon})~\cite{khosla2020supcon} objective restricted to the IoC-matched positive set:
\begin{equation}
    \mathcal{L}_{\mathrm{IoC}}(\theta; \mathcal{B}) \;=\;
    \begin{cases}
        \displaystyle -\frac{1}{W_{\mathcal{B}}}\sum_{a \in \mathcal{A}(\mathcal{B})} w_a \cdot \ell_\tau(a) & \text{if } |\mathcal{A}(\mathcal{B})| \ge 2,\\[6pt]
        0 & \text{otherwise,}
    \end{cases}
    \label{eq:contrastive_loss}
\end{equation}
where $W_{\mathcal{B}} := \sum_{a\in\mathcal{A}(\mathcal{B})} w_a$ and the per-anchor SupCon log-softmax term is
\begin{equation}
    \ell_\tau(a) \;:=\; \frac{1}{|P(a)|} \sum_{p \in P(a)} \log \frac{\exp(z_a \cdot z_p / \tau)}{\sum_{b \in \mathcal{B},\, b \neq a} \exp(z_a \cdot z_b / \tau)},
    \label{eq:supcon_anchor}
\end{equation}
with $\tau > 0$ the contrastive temperature (we use $\tau = 0.1$ in our experiments).
The denominator in $\ell_\tau(a)$ sums over all $b \in \mathcal{B}$ with $b \neq a$, so every batch sample except the anchor contributes to the denominator: non-IoC flows serve as negatives, while other IoC-matched anchors appear simultaneously as positives in the numerator. 
In the edge case $\mathcal{A}(\mathcal{B}) = \mathcal{B}$ (every batch sample is IoC-matched), no negatives remain, the objective loses its negative-repelling term and degenerates to pulling all embeddings together, so no campaign-discriminative gradient remains; $\ell_\tau(a)\to -\log(|\mathcal{B}|{-}1)$; given the sparse IoC-matched fraction reported in Sec.~\ref{sec:experiments} (1--2\%), this corner does not arise in our experiments and we flag it as a deployment caveat for IoC-dense regimes.
Ultimately, the conceptual local objective combines the two terms additively:
\begin{equation}
    \mathcal{L}_i(\theta;\mathcal{B}) = \mathcal{L}_{\mathrm{CE}}(\theta;\, \mathcal{B}) \;+\; \lambda \cdot \mathcal{L}_{\mathrm{IoC}}(\theta;\, \mathcal{B}),
    \label{eq:full_loss}
\end{equation}
where $\lambda > 0$ controls the strength of the IoC signal.
When $\mathcal{I}_i^{(t)} = \emptyset$ (no IoC available), $\mathcal{X}_i^{\mathrm{IoC}} = \emptyset$ and $\mathcal{L}_{\mathrm{IoC}} = 0$, the client-local objective reduces to cross-entropy optimization for the core use case of a classification task.

Let $g_{i,\mathrm{CE}}^{(t)} := (\theta^{(t-1)} - \theta_i^{\mathrm{CE},(t)})/\eta$ denote the pseudo-gradient~\cite{mcmahan2017fedavg} from training on $\mathcal{L}_{\mathrm{CE}}$ alone for $E$ local epochs under learning rate $\eta$ (note that for \textsc{FedProx}, $\mathcal{L}_{\mathrm{CE}}$ is replaced by the proximal-augmented objective).
Instead of minimizing Eq.~\eqref{eq:full_loss} jointly, which couples the two gradients across local SGD steps and entangles them, \textsc{FedIoC} uses a \emph{two-pass} design that defines the transmitted client update directly as
\begin{equation}
    g_i^{(t)} \;:=\; \underbrace{g_{i,\mathrm{CE}}^{(t)}}_{\text{\PhaseI{}: task component}} \;+\; \lambda \cdot \underbrace{g_{i,\mathrm{IoC}}^{(t)}}_{\text{\PhaseII{}: IoC-contrastive component}},
    \label{eq:gradient_decomp}
\end{equation}
where $g_{i,\mathrm{IoC}}^{(t)} := (\theta^{(t-1)} - \theta_i^{\mathrm{IoC},(t)})/\eta$ is computed by training on $\mathcal{L}_{\mathrm{IoC}}$ alone for one epoch starting from the \emph{same} weights $\theta^{(t-1)}$ used by \PhaseI{}.
We emphasize that $g_i^{(t)}$ is defined by Eq.~\eqref{eq:gradient_decomp}, not as the joint pseudo-gradient $(\theta^{(t-1)} - \theta_i^{(t)})/\eta$ obtained from running gradient optimization on $\mathcal{L}_{\mathrm{CE}} + \lambda \mathcal{L}_{\mathrm{IoC}}$ end-to-end; \textsc{FedIoC} is the additive construction throughout this paper.
We further note that Eq.~\eqref{eq:gradient_decomp} is a \emph{definition} of the transmitted update, not a derived decomposition: because the two passes share initial weights and are computed independently, the additive form holds by construction within a round and avoids the cross-step coupling that joint minimization of Eq.~\eqref{eq:full_loss} would induce; the cross-round interaction through $\theta^{(t)}$ is empirical and base-dependent (Sec.~\ref{sec:results}).
The additional \textbf{computational cost} is one extra local epoch per round on top of the $E$ epochs of \PhaseI{}; \textbf{communication cost} is unchanged because only the summed update $g_i^{(t)}$ is transmitted.
In expectation, $g_{i,\mathrm{IoC}}^{(t)}$ is a functional of the matched-flow distribution $P_\kappa$, so same-campaign clients produce co-directed gradients while cross-campaign clients do not.

\subsection{Server-Side Campaign Detection}
\label{subsec:server}
At the end of each round, the server holds gradients $\{g_i^{(t)}\}_{i=1}^n$ before aggregation.
Letting $\mathcal{C}^{+} = \bigl\{c_k : \|g_k^{(t)}\| > 0\bigr\}$ denote the set of non-zero-update clients (clients with $\|g_i^{(t)}\| = 0$, typically those for which $\mathcal{L}_{\mathrm{IoC}} = 0$ in every batch this round or whose \PhaseI{} update happened to vanish, are excluded as abstentions to avoid spurious singleton clusters), it computes a pairwise cosine similarity matrix $\mathbf{S} \in \mathbb{R}^{|\mathcal{C}^{+}|\times|\mathcal{C}^{+}|}$:
\begin{equation}
    S_{ij} \;=\; \frac{g_i^{(t)} \cdot g_j^{(t)}}{\|g_i^{(t)}\|\cdot \|g_j^{(t)}\|}, \qquad c_i,\, c_j \in \mathcal{C}^{+},
    \label{eq:cosine}
\end{equation}
and derives an elementwise distance matrix $D_{ij} = 1 - S_{ij}$.
Note that $1 - \cos\angle(g_i,g_j)$ is a dissimilarity but not a metric (it does not satisfy the triangle inequality in general); this is considered unproblematic here because agglomerative clustering with complete linkage operates directly on the dissimilarity matrix and does not rely on metric properties.
Agglomerative clustering with complete linkage~\cite{müllner2011modernhierarchicalagglomerativeclustering} is then applied to $\mathbf{D}$ with distance threshold $\varepsilon$, yielding the campaign partition $\mathcal{C}^{(t)} = \textsc{AgglomerativeCluster}(\mathbf{D},\, \varepsilon,\, \text{complete})$.
Here, the complete linkage property merges two clusters only when the \emph{maximum} pairwise distance between their members falls below $\varepsilon$, preventing the chain-collapse failure mode of single-linkage methods under heterogeneous gradient norms.
Each cluster groups clients whose IoC-enriched gradients are geometrically proximate, identifying a global attack campaign from gradient alignment alone.
Appendix~\ref{app:threat} details the threat model under which this no-raw-indicator-exchange property holds and enumerates the residual attack vectors against which \textsc{FedIoC} offers no defense.
The server subsequently aggregates the transmitted pseudo-gradients $g_i^{(t)}$ (Eq.~\eqref{eq:gradient_decomp}):
\begin{equation}
    \theta^{(t)} \;=\; \theta^{(t-1)} \;-\; \eta \cdot \sum_{i=1}^n \frac{|\mathcal{D}_i|}{\sum_{j=1}^n |\mathcal{D}_j|}\, g_i^{(t)}.
\end{equation}
We assume synchronous full-client participation each round, matching our experimental setup; the construction extends naturally to partial-participation schedules where the server aggregates and clusters only over the subset of clients that report in round~$t$.
The complete pseudocode of \textsc{FedIoC} is presented in Algorithm~\ref{alg:fedioc}; \colorbox{rowbase}{amber} lines mark the \PhaseI{} cross-entropy pass that any gradient-transmitting FL algorithm already performs, while \colorbox{rowours}{blue} lines mark the modular additions introduced by \textsc{FedIoC} (the \PhaseII{} IoC-contrastive pass, the additive update, and the server-side cosine clustering).
\begin{algorithm}[!h]
\caption{\small\textsc{FedIoC}: Contrastive IoC-Encoded Federated Campaign Detection}
\label{alg:fedioc}
\begin{algorithmic}[1]
\small
\Require Global model $\theta^{(0)}$, clients $\mathcal{C} = \{c_1,\ldots,c_n\}$, rounds $T$,
         local epochs $E$ (we use $E=1$),
         $\lambda$, $\tau$, distance threshold $\varepsilon$, local learning rate $\eta$
\Ensure  Final model $\theta^{(T)}$, campaign clusters $\{\mathcal{C}^{(t)}\}_{t=1}^T$
\For{$t = 1$ \textbf{to} $T$}
    \State \textbf{Server} broadcasts $\theta^{(t-1)}$ to all clients
    \ForEach{client $c_i \in \mathcal{C}$ \textbf{(in parallel)}}
        \State Retrieve IoC set $\mathcal{I}_i^{(t)}$ from local CTI feed
        \HlBase{\textbf{\PhaseI{}: CE gradient} (classification, $E$ local epochs from $\theta^{(t-1)}$):}
        \HlBase{$\theta_i^{\mathrm{CE}} \leftarrow \theta^{(t-1)}$}
        \For{$e = 1$ \textbf{to} $E$}
            \ForEach{mini-batch $\mathcal{B} \subseteq \mathcal{D}_i$}
                \HlBase{$\theta_i^{\mathrm{CE}} \leftarrow \theta_i^{\mathrm{CE}} - \eta\,\nabla_\theta \mathcal{L}_{\mathrm{CE}}(\theta_i^{\mathrm{CE}}; \mathcal{B})$}
            \EndFor
        \EndFor
        \HlBase{$g_{i,\mathrm{CE}}^{(t)} \leftarrow \bigl(\theta^{(t-1)} - \theta_i^{\mathrm{CE}}\bigr)/\eta$}
        \If{$\mathcal{I}_i^{(t)} \neq \emptyset$}
            \State Match local samples: $\mathcal{X}_i^{\mathrm{IoC}} \leftarrow \{x \in \mathcal{D}_i \mid \exists\, \iota_k \in \mathcal{I}_i^{(t)}: x \models \pi_k\}$ \Comment{Eq.~\eqref{eq:ioc_match}}
            \HlOurs{\textbf{\PhaseII{}: IoC gradient} (campaign encoding module):}
            \HlOurs{$\theta_i^{\mathrm{IoC}} \leftarrow \theta^{(t-1)}$}
            \ForEach{mini-batch $\mathcal{B} \subseteq \mathcal{D}_i$ with $|\mathcal{A}(\mathcal{B})| \geq 2$} \Comment{Eq.~\eqref{eq:contrastive_loss}}
                \HlOurs{$\theta_i^{\mathrm{IoC}} \leftarrow \theta_i^{\mathrm{IoC}} - \eta\,\nabla_\theta \mathcal{L}_{\mathrm{IoC}}(\theta_i^{\mathrm{IoC}}; \mathcal{B})$}
            \EndFor
            \HlOurs{$g_{i,\mathrm{IoC}}^{(t)} \leftarrow \bigl(\theta^{(t-1)} - \theta_i^{\mathrm{IoC}}\bigr)/\eta$}
            \HlOurs{$g_i^{(t)} \leftarrow g_{i,\mathrm{CE}}^{(t)} + \lambda \cdot g_{i,\mathrm{IoC}}^{(t)}$ \Comment{Two-pass additive update (Eq.~\ref{eq:gradient_decomp})}}
        \Else
            \State $g_i^{(t)} \leftarrow g_{i,\mathrm{CE}}^{(t)}$ \Comment{Falls back to the underlying aggregation rule}
        \EndIf
        \State Transmit $g_i^{(t)}$ to server
    \EndFor
    \State \textbf{Server} computes $S_{ij} \leftarrow \cos\bigl(g_i^{(t)}, g_j^{(t)}\bigr)$ for all $i,j$; derives $D_{ij} \leftarrow 1 - S_{ij}$
    \HlOurs{\textbf{Server} detects campaigns: $\mathcal{C}^{(t)} \leftarrow \textsc{AgglomerativeCluster}(\mathbf{D},\, \varepsilon,\, \text{complete})$}
    \State \textbf{Server} aggregates: $\theta^{(t)} \leftarrow \theta^{(t-1)} - \eta \cdot \sum_{i=1}^n \tfrac{|\mathcal{D}_i|}{\sum_j|\mathcal{D}_j|}\, g_i^{(t)}$
\EndFor
\State \Return $\theta^{(T)},\; \{\mathcal{C}^{(t)}\}_{t=1}^T$
\end{algorithmic}
\end{algorithm}

\subsection{Compatibility with FL Methods}
\label{subsec:compat}

\textsc{FedIoC} has two orthogonal components: the client-side IoC objective (Eq.~\eqref{eq:full_loss}) is agnostic to the server's aggregation rule, and the server-side clustering (Sec.~\ref{subsec:server}) is agnostic to the client training algorithm.
The single structural requirement is that the server observes individual client updates to compute pairwise cosine similarities, which is satisfied by \textsc{FedAvg}~\cite{mcmahan2017fedavg}, \textsc{FedProx}~\cite{li2020fedprox}, and \textsc{SCAFFOLD}~\cite{karimireddy2020scaffold} but not by secure aggregation~\cite{bonawitz2017secagg} or split/vertical FL~\cite{khan2025vfl}; this is a fundamental privacy-utility trade-off, since secure aggregation is the canonical mitigation against the gradient-inversion channel discussed in App.~\ref{app:threat}.
Protocol-level compatibility does not guarantee effective clustering either: algorithms that strongly homogenize inter-client gradient directions (e.g.\ \textsc{FedProx} with large $\mu$) would suppress the very diversity the clustering exploits.


\section{Experimental Setup}
\label{sec:experiments}

\subsection{Datasets and IoC Extraction}

\noindent\textbf{CTU-13}~\cite{garcia2014empirical} contains real botnet traffic from 13 distinct campaigns captured on a university network, each campaign corresponding to a different botnet family.
The structured campaign labeling makes CTU-13 our primary benchmark: ground-truth campaign membership provides an unambiguous reference partition for computing cluster quality metrics, and the shared command-and-control infrastructure within each family yields the cohesive IoC sets that the contrastive objective is designed to exploit.

\noindent\textbf{UNSW-NB15}~\cite{moustafa2015unswnb15} contains nine heterogeneous attack families (treated as campaigns) that span reconnaissance, exploits, fuzzers, DoS, worms, generic, backdoors, analysis, and shellcode.
This diverse mix lacks the shared C\&C infrastructure binding flows within a single botnet family, and we use UNSW-NB15 as a generalization test for whether the gradient-clustering signal extends beyond cohesive infrastructure-bound campaigns.

\noindent\textbf{IoC extraction.}
For each client $c_i$ we extract source IP addresses from malicious flows in $\mathcal{D}_i$ and encode them as STIX \texttt{Indicator} objects, simulating clients deriving indicators from observed network traffic.
The STIX \texttt{confidence} field is set to $\beta_k = \log(1 + n_k)/\log(1 + \max_{k'} n_{k'})$, where $n_k$ counts the matched malicious flows in the issuing client's partition, so frequently-matched indicators receive weight near $1$ and one-off matches a fractional weight via the per-anchor weight $w_a$ (when $\max_{k'} n_{k'} = 0$ no IoC are matched and $\mathcal{L}_{\mathrm{IoC}} = 0$, so the formula is vacuous).
All indicators are available from round~1, and the IoC-matched fraction is sparse on both datasets (1--2\% of flows on CTU-13), so most mini-batches satisfy $|\mathcal{A}(\mathcal{B})| < 2$ and $\mathcal{L}_{\mathrm{IoC}} = 0$.


\subsection{Federated Setup}

Ten clients are constructed via a campaign-stratified non-IID partition of CTU-13: each scenario is split into three equal chunks and assigned round-robin so every client sees 3--5 campaigns but never a complete view of any single campaign, mirroring a realistic scenario in which organizations observe different segments of attacker infrastructure.
Each client therefore holds \emph{partial} IoC, with STIX indicators covering only the campaign fragments present in its local slice.
UNSW-NB15 uses the identical partitioning strategy and pipeline without dataset-specific tuning.
Training runs for 15 FL rounds with a small MLP over five flow-level features.
Implementation details are outlined in Appendix~\ref{app:setup}.
Figure~\ref{fig:attribution} further visually demonstrates the clustering spectrum that the FL server produces from the per-client gradient updates.
\begin{figure}[h!]
    \centering
    \includegraphics[width=1.0\linewidth]{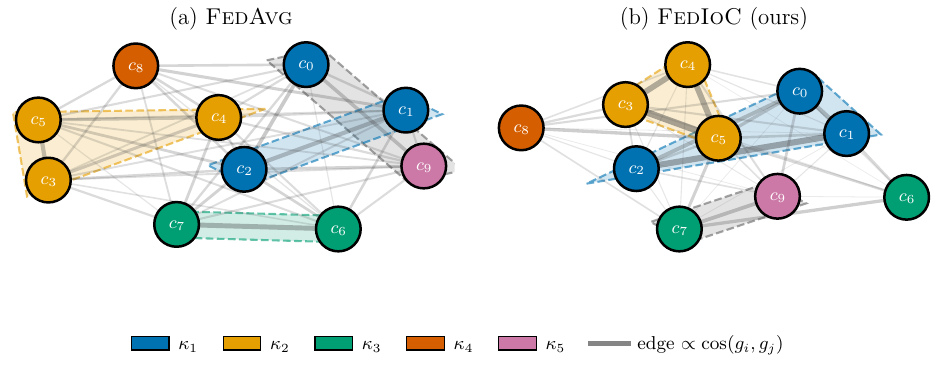}
    \caption{Single-round visualization of the FL server's campaign-cohort output computed from the uploaded per-client gradient updates under (a) \textsc{FedAvg} and (b) \textsc{FedIoC} on CTU-13. Nodes are federated clients colored by their dominant ground-truth campaign $\kappa_k$; edge opacity scales with the similarity $\cos\bigl(g_i, g_j\bigr)$ of their gradient updates; dashed hulls enclose the clusters returned by the server's agglomerative clustering step at threshold $\varepsilon=0.5$ (gray hull = multi-campaign cluster). \textbf{Takeaway:} the server recovers campaign-aligned clusters directly from gradient geometry; in this illustrative round the IoC-contrastive variant (b) yields tighter single-campaign hulls than \textsc{FedAvg} (a).}
    \label{fig:attribution}
\end{figure}

\subsection{Method Baselines}

\begin{itemize}
    \item Local-only: a single representative client trains on its local partition without federation. Confirms that campaign-cohort recovery requires cross-client gradient exchange.
    \item \textsc{FedAvg}~\cite{mcmahan2017fedavg}: standard federated averaging with no IoC. Measures the campaign structure plain gradient clustering recovers from non-IID partitions.
    \item \textsc{FedProx}~\cite{li2020fedprox}: \textsc{FedAvg} with a proximal regularization term limiting per-client drift.
    \item \textsc{SCAFFOLD}~\cite{karimireddy2020scaffold}: variance-reduced FL using server- and client-side control variates to counter client drift.
    \item \textsc{FedProx}+IoC and \textsc{SCAFFOLD}+IoC: plugin demonstrations combining the IoC contrastive loss (Sec.~\ref{subsec:compat}) with each base method, testing whether the contrastive signal composes with proximal + variance-reduction regularizers.
    \item \textsc{FedAvg}+SupCon-lbl: an ablation control identical to \textsc{FedIoC} except the contrastive positive set is the malicious class label instead of IoC matches; it isolates whether the indicator-specific objective contributes beyond contrastive up-weighting of the minority class.
    \item \textsc{FedIoC} (ours): IoC-contrastive gradients over a \textsc{FedAvg} base with server-side agglomerative campaign detection.
\end{itemize}

\subsection{Evaluation Metrics}

\paragraph{Campaign recovery (primary):}
Adjusted Rand Index~(\textbf{ARI})~\cite{hubert1985ari} and Normalized Mutual Information~(\textbf{NMI})~\cite{strehl2002nmi} between cluster assignments $\mathcal{C}^{(t)}$ and ground-truth campaign labels.
We further report both \emph{peak ARI} ($\max_{t} \mathrm{ARI}^{(t)}$) and \emph{mean ARI} over an early detection window. 

\paragraph{Detection performance (secondary):}
Macro-averaged \textbf{F1}~\cite{sokolova2009measures} on the held-out test set, confirming that IoC encoding does not degrade intrusion detection capability.


\section{Preliminary Results and Discussion}
\label{sec:results}
\subsection{Campaign Detection}
Table~\ref{tab:main} reports multi-seed campaign-recovery and classification metrics across all methods in both cyber attack scenarios.
Our main findings are as follows:

\noindent\textbf{Gradient geometry recovers campaign cohorts, but the IoC encoding needs further investigation.}
Across both benchmarks the FL server recovers campaign cohorts directly from the cosine geometry of client gradients: on CTU-13 every contrastive variant and \textsc{FedAvg} reach high agreement with the ground truth (Peak ARI $0.89$--$0.97$, Mean ARI $0.76$--$0.86$; Fig.~\ref{fig:attribution}).
This recovery is the capability \textsc{FedIoC} targets, but it is not exclusively attributable to the indicator-specific objective: on CTU-13 \textsc{FedIoC} (Peak $0.97$, Mean $0.85$) lies within run-to-run variance of a label-only contrastive control (\textsc{FedAvg+SupCon-lbl}, $0.94$/$0.86$) and of plain \textsc{FedAvg} ($0.97$/$0.80$) (overlapping $\pm1$\,SD, Table~\ref{tab:main}), on UNSW-NB15 \textsc{FedAvg}/\textsc{FedProx} match or exceed it, and any contrastive advantage appears only at larger learning rates that suppress the \textsc{FedAvg} baseline.
The indicator-weighted objective therefore does not yet yield a gain separable from contrastive up-weighting of the malicious class; isolating a regime in which IoC-identity encoding provably helps is the central open problem.
\begin{table}[b!]
\centering
\caption{Method comparison on CTU-13 (13 botnet campaigns) and UNSW-NB15 (9 attack-family campaigns), mean\,$\pm$\,SD across 3 seeds ($E{=}1$, learning rate~$10^{-4}$). \textbf{Takeaway}: the FL server recovers campaign cohorts from gradient geometry across methods, but on CTU-13 the contrastive variants are statistically comparable to \textsc{FedAvg} and to the label-only control. \textbf{Bold}: within 1\,SD of the best CTU-13 value per column; UNSW-NB15 entries are comparable within variance and left unmarked.}
\label{tab:main}
\resizebox{\textwidth}{!}{%
\begin{tabular}{lcccccccc}
\toprule
& \multicolumn{2}{c}{F1} & \multicolumn{2}{c}{Peak ARI} & \multicolumn{2}{c}{Mean ARI (R1--7)} & \multicolumn{2}{c}{NMI} \\
\cmidrule(lr){2-3} \cmidrule(lr){4-5} \cmidrule(lr){6-7} \cmidrule(lr){8-9}
Method & CTU-13 & UNSW & CTU-13 & UNSW & CTU-13 & UNSW & CTU-13 & UNSW \\
\midrule
\rowcolor{rowbase}Local-only       & $0.48\pm0.01$ & $0.42\pm0.26$ & -- & -- & -- & -- & -- & -- \\
\rowcolor{rowbase}\textsc{FedAvg}           & $0.49\pm0.01$ & $0.48\pm0.09$ & $\mathbf{0.97\pm0.05}$ & $0.58\pm0.33$ & $\mathbf{0.80\pm0.16}$ & $0.47\pm0.33$ & $\mathbf{0.95\pm0.05}$ & $0.84\pm0.08$ \\
\rowcolor{rowbase}\textsc{FedProx}          & $0.49\pm0.00$ & $0.49\pm0.10$ & $0.89\pm0.09$ & $0.58\pm0.33$ & $0.69\pm0.17$ & $0.49\pm0.30$ & $0.88\pm0.06$ & $0.84\pm0.08$ \\
\rowcolor{rowbase}\textsc{SCAFFOLD}         & $0.36\pm0.19$ & $0.23\pm0.20$ & $0.59\pm0.13$ & $0.42\pm0.37$ & $0.42\pm0.18$ & $0.06\pm0.05$ & $0.81\pm0.06$ & $0.82\pm0.03$ \\
\midrule
\rowcolor{rowours}\textsc{FedProx}+IoC      & $0.49\pm0.01$ & $0.45\pm0.08$ & $0.89\pm0.09$ & $0.58\pm0.33$ & $\mathbf{0.76\pm0.16}$ & $0.43\pm0.29$ & $\mathbf{0.93\pm0.07}$ & $0.89\pm0.09$ \\
\rowcolor{rowours}\textsc{SCAFFOLD}+IoC     & $0.28\pm0.16$ & $0.23\pm0.20$ & $0.72\pm0.09$ & $0.33\pm0.37$ & $0.36\pm0.23$ & $0.05\pm0.05$ & $0.79\pm0.10$ & $0.82\pm0.03$ \\
\rowcolor{rowours}\textsc{FedAvg}+SupCon-lbl & $0.50\pm0.01$ & $0.43\pm0.09$ & $\mathbf{0.94\pm0.10}$ & $0.53\pm0.27$ & $\mathbf{0.86\pm0.16}$ & $0.39\pm0.24$ & $\mathbf{0.93\pm0.07}$ & $0.86\pm0.06$ \\
\rowcolor{rowours}\textsc{FedIoC} (ours) & $0.50\pm0.01$ & $0.45\pm0.08$ & $\mathbf{0.97\pm0.05}$ & $0.58\pm0.33$ & $\mathbf{0.85\pm0.14}$ & $0.43\pm0.32$ & $\mathbf{0.94\pm0.05}$ & $0.89\pm0.09$ \\
\bottomrule
\end{tabular}%
}
\end{table}


\noindent\textbf{Composition with regularized bases is uneven.}
Adding the contrastive term to \textsc{FedProx} improves its CTU-13 Mean ARI ($0.69\to0.76$) at no F1 cost, whereas on \textsc{SCAFFOLD} it raises Peak ARI ($0.59\to0.72$) but degrades Mean ARI and F1; \textsc{SCAFFOLD} is moreover unstable in our $E{=}1$ regime (F1~$0.36$, high variance).
The contrastive signal thus composes unevenly with proximal and variance-reduction regularizers.

\noindent\textbf{Gradient clustering recovers campaign cohorts without raw IoC exchange.}
Each cluster is a set of clients whose gradient updates are geometrically proximate, which our results tie to a shared campaign-correlated traffic distribution, not to the indicator-specific component alone.
The server's cluster output is thus an implicit campaign-cohort report: it identifies which organizations observe the same attacker infrastructure without any organization disclosing which specific indicators it holds (raw-indicator exposure only; cohort membership and gradient-inversion channels are discussed in App.~\ref{app:threat}).
Detection concentrates in early rounds: client gradients are most campaign-discriminative while clients are still diverging from a common initialization, and as the shared model converges they grow more homogeneous and the campaign signal weakens.
Practically, this matches the operational lifecycle of IP-based IoC~\cite{liao2016acing}, for which indicators are most actionable immediately after issuance.

\subsection{Classification Performance}

The two-pass design (Sec.~\ref{subsec:weighting}) yields approximate \emph{non-interference}: the IoC pass leaves classification essentially unchanged, with \textsc{FedIoC}'s macro-F1 within noise of \textsc{FedAvg} on both datasets (CTU-13 $0.50$ vs $0.49$; UNSW-NB15 $0.45$ vs $0.48$), and the label-only control matching it as well ($0.50$ / $0.43$).
Absolute F1 is modest across all non-degenerate methods ($0.43$--$0.50$) at this configuration, which favors gradient-clustering stability ($E{=}1$, lr~$10^{-4}$) over classifier fitting; the \textsc{SCAFFOLD} variants are degenerate here (F1~$0.23$--$0.36$, high variance).
So while the IoC objective does not \emph{degrade} detection, it also does not improve it, mirroring the clustering result.
Class imbalance (2.2\% botnet traffic in CTU-13) is handled by class-weighted cross-entropy on all training flows; the contrastive loss operates over penultimate embeddings without separate class weighting.

\section{Related Work}
\label{sec:related_work}

\paragraph{FL for Threat Detection and Gradient-Level Encoding.}
FL has been applied extensively to NIDS and IoT anomaly detection~\cite{campos2022fl}; surveys of federated cyber intelligence~\cite{tabrizchi2025fedcyber} confirm that existing systems treat IoC exclusively as training-time artifacts, so the gradient is a statistical artifact of local loss minimization and not a vehicle for encoded threat knowledge.
Dataset condensation via gradient matching~\cite{zhao2021dataset} establishes that gradients can be sculpted to carry specific semantic knowledge, the key property \textsc{FedIoC} exploits: same-campaign clients produce contrastively-aligned gradients the server can cluster without observing any raw indicator.

\paragraph{FL Aggregation under Heterogeneity.}
\textsc{FedAvg}~\cite{mcmahan2017fedavg}, \textsc{FedProx}~\cite{li2020fedprox}, and \textsc{SCAFFOLD}~\cite{karimireddy2020scaffold} are the standard FL aggregation methods for non-IID settings; \textsc{FedProx} and \textsc{SCAFFOLD} specifically aim to suppress inter-client gradient divergence via proximal regularization and variance reduction with control variates, respectively.
\textsc{FedIoC} takes the opposite stance: instead of suppressing gradient divergence, it reads the divergence already present under non-IID partitions as a campaign signal and clusters on it.
In our experiments the contrastive term composes acceptably with proximal regularization (\textsc{FedProx}) but interacts poorly with variance reduction (\textsc{SCAFFOLD}), which is unstable in our regime; characterizing these interactions is left open.

\paragraph{Supervised Contrastive Learning and FL.}
Supervised contrastive learning~\cite{khosla2020supcon} extends the self-supervised \textsc{SimCLR} framework~\cite{chen2020simclr} to the labeled setting: same-class samples form the positive set, and the loss maximizes their embedding similarity relative to all other batch samples.
Bringing this objective into federated learning, the closest prior work is \textsc{MOON}~\cite{li2021moon}, which contrasts each client's local representation against the global model to curb non-IID client drift.
\textsc{FedIoC} inverts this: instead of contrasting against the global model to \emph{suppress} drift, it contrasts indicator-matched flows \emph{within} each client so that same-campaign clients produce aligned gradient directions, repurposing the supervised contrastive signal from a drift regularizer into a server-readable campaign-coordination primitive.

\paragraph{Research Gap.}
All three threads miss the same opportunity: none uses the client gradient to carry threat-indicator structure that a server can read to recover campaigns across organizations.
Federated threat-detection systems treat IoC as training labels and stop at per-flow classification~\cite{tabrizchi2025fedcyber,campos2022fl}.
Aggregation methods treat inter-client gradient divergence as instability to suppress, not as a signal.
And supervised contrastive learning shapes embeddings inside one model, never to align gradients across clients.
\textsc{FedIoC} fills this gap: it encodes indicators contrastively so that each client's gradient signals its campaign membership, and the server clusters these gradients to recover the global campaign partition.

\section{Conclusion}
\label{sec:conclusion}

We propose a modular framework for gradient-space campaign attribution that gives the main research question of Section~\ref{ref:rq} a \emph{partial, affirmative} answer:
the server detects campaign cohorts by cosine clustering over the uploaded gradients, without raw indicator transmission.
However, our controlled study shows this recovery arises largely from the non-IID gradient structure, since the indicator-weighted objective is not separable from a label-only control on the benchmark dataset; we therefore offer \textsc{FedIoC} as an initial infrastructure for follow-up investigation.
\textbf{Future work:} The highest-priority target is evaluating more effective gradient-encoding methods within our novel framework.
Further directions include whether the encoding channel is \emph{task-agnostic} (the cohort signal surviving when clients optimize an unrelated primary task such as image classification, making gradient-level coordination general-purpose), streaming-campaign protocols in which indicators arrive mid-federation, and extension to richer cyber observables such as file hashes or registry key changes.
\begin{credits}
\subsubsection{\ackname}
This work is funded by the European Regional Development Fund (ERDF) under grant FKZ: 2404-003-1.2 (EU-EFRE GREEN-INNO), as well as by ProPere THWS, the Center for Cybersecurity TTZ-WUE, and the Center for Artificial Intelligence Würzburg (CAIRO).

\end{credits}

\appendix

\bibliographystyle{splncs04}
\bibliography{references}


\clearpage
\section*{Appendix}

\renewcommand{\theHsection}{\Alph{section}}

\section{Implementation Details}
\label{app:setup}

\noindent\textbf{Model and optimization.}
The global model is a three-hidden-layer MLP (input $\to$ 256 $\to$ 128 $\to$ 64 $\to$ output) with ReLU activations and dropout 0.3, operating on five flow-level features: duration, total packets, total bytes, source bytes, and protocol (encoded as an integer).
Training uses the Adam optimizer with learning rate $1\times 10^{-4}$, batch size 256, and $E=1$ local epoch per round.
$E=1$ is deliberate: additional local epochs homogenize client gradients and collapse the cosine-similarity structure the server clusters on (CTU-13 Peak ARI falls from $0.97$ at $E{=}1$ to below $0.40$ at $E{=}10$).
The clustering signal is likewise sensitive to the local learning rate: at $10^{-4}$ plain \textsc{FedAvg} already recovers campaigns well, whereas a larger rate ($10^{-3}$) suppresses the baseline and inflates the apparent benefit of the contrastive objective, which is why we report the smaller, more conservative rate.
The reported experiments use \texttt{sample\_frac}$=0.1$ (10\% of each scenario's rows) to enable rapid iteration.

\noindent\textbf{Hyperparameters.}
We use $\lambda{=}1.0$ on both datasets; the clustering threshold is $\varepsilon{=}0.5$ on CTU-13 and $\varepsilon{=}0.07$ on UNSW-NB15.
The contrastive temperature is $\tau{=}0.1$ throughout, and \textsc{FedProx} uses the literature-standard $\mu{=}0.01$.
Because $g_{i,\mathrm{IoC}}^{(t)} = (\theta^{(t-1)} - \theta_i^{\mathrm{IoC}})/\eta$ accumulates one full \PhaseII{} epoch of mini-batch SGD steps, the effective contrastive strength scales with the number of IoC-bearing batches per client; the headline $\lambda{=}1.0$ is therefore not directly portable across datasets with very different IoC-match counts, and a $\lambda$ rescaled by \PhaseII{} step count (or an explicit server learning rate separate from $\eta$) is the natural reformulation for cross-dataset transfer.
Other baseline hyperparameters are held at standard literature values.

\noindent\textbf{Detection window.}
The 7-round window for $\bar{\mathrm{ARI}}$ is motivated operationally by the short actionable lifetime of IP-based IoC~\cite{liao2016acing}: source-IP indicators typically remain useful for hours to a few days before attacker infrastructure rotation degrades coverage, corresponding to the earliest federation rounds under any realistic round cadence.
We complement the early-window mean with Peak ARI throughout to make the round at which each method's signal is strongest visible to the reader, and report all-rounds curves in the per-round figures so the early-window choice does not hide late-round behaviour.

\section{Threat Model and Privacy Properties}
\label{app:threat}

\textsc{FedIoC} assumes an \emph{honest-but-curious} server and honest clients.
The mechanism guarantees one concrete property: raw IoC patterns never leave the client; only the blended gradient $g_i^{(t)}$ is transmitted, and when $\mathcal{I}_i^{(t)}=\emptyset$ this reduces to a standard \textsc{FedAvg} update.
\textsc{FedIoC} does \emph{not} defend against (i)~gradient inversion attacks~\cite{geiping2020inverting}, (ii)~malicious clients injecting adversarial indicators (analogous to FL data poisoning~\cite{nguyen2022flame}), or (iii)~a compromised server using the cohort report offensively; the cohort graph itself is a sensitive artifact whose governance lies outside the protocol guarantees of \textsc{FedIoC}.
Gradient inversion is a particular concern because the IoC-contrastive component $\lambda g_{i,\mathrm{IoC}}^{(t)}$ encodes IoC-membership by design, so inversion targets the matched source IPs more directly than the CE component alone, and per-round observation of $g_i^{(t)}$ compounds across rounds as a separate leakage channel.
Secure aggregation~\cite{bonawitz2017secagg} is structurally incompatible with the per-client visibility our clustering requires, so the relevant mitigation is per-client differential privacy on the IoC component instead of aggregation-time hiding; \textsc{FedIoC} provides no formal differential-privacy guarantee, and bounding $I(g_i^{(t)};\,\mathcal{I}_i^{(t)})$ together with applying DP~\cite{abadi2016deep} to the IoC gradient component are left to future work.

\end{document}